\documentclass[pmlr]{jmlr}

\usepackage{longtable}

\usepackage{booktabs}
\usepackage[load-configurations=version-1]{siunitx} 

\theorembodyfont{\upshape}
\theoremheaderfont{\scshape}
\theorempostheader{:}
\theoremsep{\newline}

\jmlrvolume{85}
\jmlryear{2019}
\jmlrworkshop{Machine Learning for Healthcare}

\firstpageno{1}

\title[Temporal Graph Convolutional Networks]{Temporal Graph Convolutional Networks for Automatic Seizure Detection}

\usepackage{multirow}
\usepackage{bbm}
\usepackage{makecell}
\newcommand{\reals}{\mathbbm{R}}

\author{\Name{Ian C. Covert} \Email{icovert@cs.washington.edu}
      \addr \\
      University of Washington\\
      Seattle, Washington, USA
      \AND
      \Name{Balu Krishnan} \Email{krishnb@ccf.org} \\
      \Name{Imad Najm} \Email{najmi@ccf.org}
      \addr \\
      Cleveland Clinic Foundation\\
      Cleveland, Ohio, USA
      \AND
      \Name{Jiening Zhan} \Email{jiening@google.com} \\
      \Name{Matthew Shore} \Email{mashore@google.com} \\
      \Name{John Hixson} \Email{jhixson@google.com} \\
      \Name{Ming Jack Po} \Email{mingp@google.com}
      \addr \\
      Google AI Healthcare\\
      Mountain View, California, USA}

\begin{document}

\maketitle

\begin{abstract}
    Seizure detection from EEGs is a challenging and time consuming clinical problem that would benefit from the development of automated algorithms.
    EEGs can be viewed as \textit{structural time series}, because they are multivariate time series where the placement of leads on a patient's scalp provides prior information about the structure of interactions.
    Commonly used deep learning models for time series don't offer a way to leverage structural information, but this would be desirable in a model for structural time series.
    To address this challenge, we propose the temporal graph convolutional network (TGCN), a model that leverages structural information and has relatively few parameters. TGCNs apply feature extraction operations that are localized and shared over both time and space, thereby providing a useful inductive bias in tasks where one expects similar features to be discriminative across the different sequences.
    In our experiments we focus on metrics that are most important to seizure detection, and demonstrate that TGCN matches the performance of related models that have been shown to be state-of-the-art in other tasks.
    Additionally, we investigate interpretability advantages of TGCN by exploring approaches for helping clinicians determine when precisely seizures occur, and the parts of the brain that are most involved.
\end{abstract}

\section{Introduction}

Epilepsy is a chronic neurological condition characterized by abnormal brain activity that leads to seizures, periods of unusual behaviors and/or sensations, and sometimes even loss of consciousness. Due to the unpredictable timing and nature of epileptic seizures, epilepsy has an outsized impact on the quality of life of its patients. The definitive diagnosis and the choice of therapy for epilepsy is based on the expert analysis of electroencephalograms (EEGs), high frequency recordings of electrical brain activity measured via 20+ electrodes placed on a patient's scalp, collected over hours to days. At 200Hz, this results in billions of data points that must be manually inspected and evaluated by neurologists.

In fact, neurologists go through additional multi-year fellowship training in the reading of EEGs to properly evaluate these signals. Even with the additional training, it takes EEG neurologists two to three hours to evaluate just a single patient's daily EEG records, creating a significant bottleneck in the diagnosis and treatment of patients.

Automated systems for detecting seizures and analyzing EEG signals could help significantly alleviate this bottleneck. A sufficiently accurate seizure detection system could aid providers in several ways: by screening and potentially characterizing EEGs that are likely to contain seizures, by integrating into an assisted reading tool for clinicians, or by performing real-time seizure detection in settings such as the intensive care unit (ICU) where an EEG expert is usually not available.

Much effort has gone into the development of seizure detection algorithms in the past decade \citep{ahammad2014detection, furbass2015prospective, golmohammadi2017gated, golmohammadi2017deep, shoeb2010application, thodoroff2016learning, williamson2012seizure}. Traditional approaches leverage signal processing techniques in combination with hand-engineered features such as entropy, Fourier transform coefficients, and wavelets \citep{ahammad2014detection, furbass2015prospective, shoeb2010application, williamson2012seizure, wilson2004seizure}. Deep learning represents a promising alternative to traditional approaches because it removes the need for hand-crafted features, and offers the potential to perform better by learning from larger datasets.

\begin{figure}[!t]
  \centering 
  \includegraphics[width=0.8\textwidth]{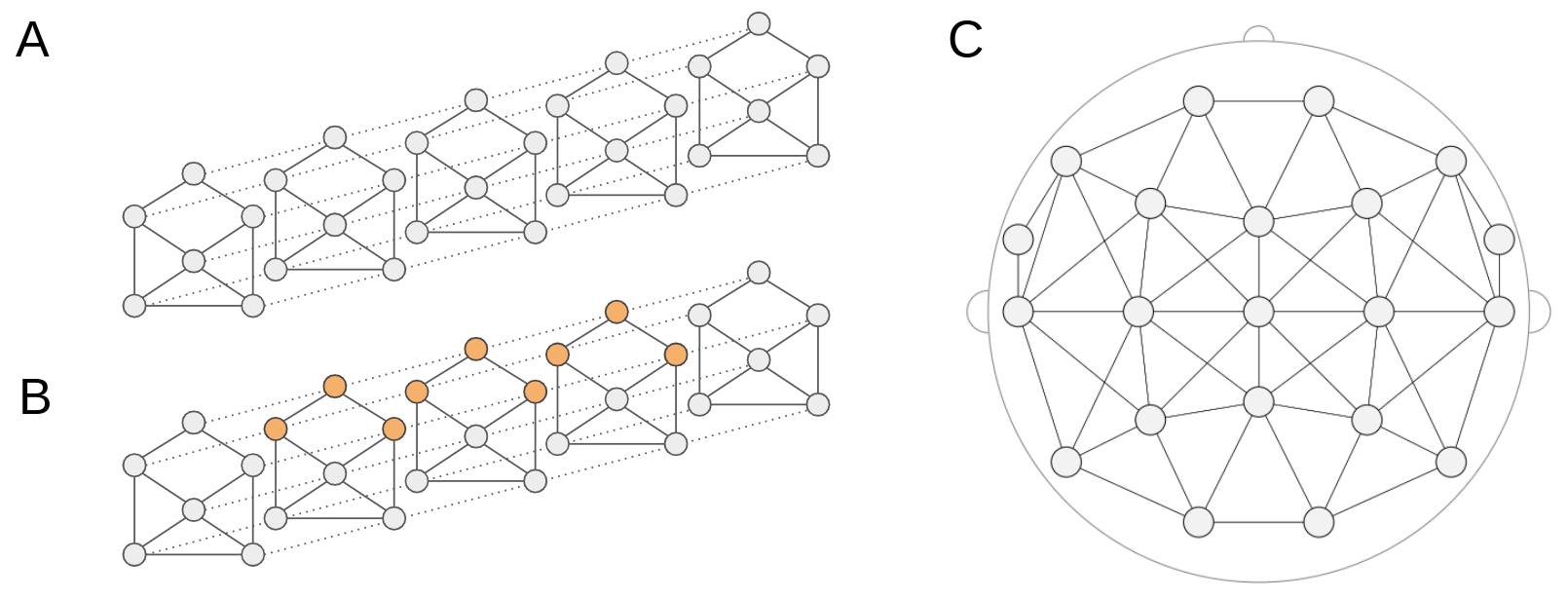}
  \caption{Structural time series. \textbf{A:} example of a structural time series with $p = 6$ sequences. Solid lines depict graph structure, dashed lines indicate temporal adjacency. \textbf{B:} depiction of the receptive field of a feature extraction operation in TGCN. TGCN applies operations that are spatially and temporally localized. In this case, the top node's neighborhood is being processed. \textbf{C:} graph topology of the $p = 21$ EEG leads in our dataset.}
  \label{fig:sts} 
\end{figure}

However, one aspect of EEGs that makes them difficult to analyze with deep learning is the structure of leads placed on a patient's scalp; this property is not straightforward to leverage with common model architectures. EEGs are an example of a \textit{structural time series} (Figure \ref{fig:sts}), which have the form $(X, A)$ where $X \in \reals^{T \times p \times c}$ contains $c$-dimensional observations across $T$ time steps for $p$ sequences, and $A \in \{0, 1\}^{p \times p}$ represents the graph topology of the $p$ sequences. (In the remainder of the paper, we refer to the different time series, which are nodes in the graph, as either nodes or sequences.) Structural time series arise in numerous applications besides EEG, including economics, traffic analysis, and human kinematics. The structural information represented by $A$ is a form of prior knowledge, and an ideal model would leverage this information to learn more efficiently and generalize better. Few deep learning models can do so, and this provides the motivation for our work.

Convolutional models have been very successful on a variety of time series tasks \citep{bai2018empirical, oord2016wavenet, wang2017time}. One particularly successful model type is the temporal convolutional neural network (TCNN) \citep{lea2017temporal}. This model applies 1D convolutions at each layer, similarly to how standard 2D CNNs process image data, and is sometimes augmnented with residual connections and dilated convolutions. Recent studies have shown that this straightforward model architecture is highly effective on a wide variety of time series classification tasks \citep{wang2017time}, and that it's competitive with recurrent neural networks (RNNs) even on tasks where an unbounded receptive field was thought to be crucial \citep{bai2018empirical}. However, as we explain concretely in the related work section, TCNN is not designed to leverage structure.

In this work we propose TGCN, a deep learning model that leverages spatial information in structural time series. In order to properly utilize the graph topology of structural time series, TGCN applies feature extractors that are localized and shared over both the temporal and spatial dimensions of the input (Figure \ref{fig:sts}B). TGCNs thus have a built-in invariance to when and where patterns occur, which is a useful inductive bias for applications where similar features are discriminative when observed in different sequences. We show that the ability of this architecture to leverage spatial information leads to performance on par with models that are state-of-the-art in other tasks. Additionally, we investigate methods for explaining TGCN predictions; in particular, we use the properties of TGCN to produce rich visualizations that shed light on when precisely, and where in the brain seizures occurred.

\section{Temporal graph convolutional networks}

\subsection{The TGCN model}

TGCN is a model that takes structural time series as input. A structural time series is represented as $(X, A)$ where $X \in \reals^{T \times p \times c}$ is a multivariate time series ($T$ is the number of time steps, $p$ is the number of sequences, and $c$ is the number of channels), and $A \in \{0, 1\}^{p \times p}$ is an adjacency matrix. $A$ can be symmetric for undirected graphs, or asymmetric for directed graphs, and has 1's on its diagonal.

At layer $l$, TGCN computes a hidden representation $h^l \in \reals^{T^l \times p \times c^l}$ in a hierarchical manner by the composition of multiple spatio-temporal convolutional (STC) layers, which are described below. Hidden representations contain features for each of the input sequences, so that the graph topology of the input is maintained at each layer. After several STC layers, the model can produce node-level predictions by outputting $h^L$, the representation at the final layer. Alternatively, it can produce single outputs (scalar or vector) by aggregating information over $h^L$ by a combination of a spatial aggregation operation (e.g. mean, max) and additional fully connected layers.

The main innovation of TGCN is that it uses feature extraction operations that are shared over both time and space. The motivation for this design is to reduce the number of parameters, prevent overfitting, make gradient-based learning easier, and enforce useful invariances. It should be noted that localized and shared feature extraction was at the core of CNNs when they were introduced for computer vision \citep{lecun1999object}. Replicating the behavior of standard 2D CNNs on lattice-like graphs (i.e. images) is difficult in the context of structural time series because of the arbitrary graph topology. In this respect, TGCN is unique among time series models (Figure \ref{fig:concept}).

\subsection{Spatio-temporal convolutional layers} \label{sec:math}

STC layers perform localized feature extraction that is shared over both the temporal and spatial dimensions of the input. This is an analogue of 2D convolutions over lattice-like graphs for structural time series. A key challenge for designing such operations is the different number of neighbors at each node, due to the arbitrary graph topology of structural time series. Graph neural networks (GNNs) manage a similar obstacle by using neighborhood aggregation schemes \citep{kipf2016semi, hamilton2017inductive, xu2018powerful}, i.e. by operating on each node independently and then aggregating the results across each node's neighborhood. Similarly, STC layers operate on each sequence independently, and then aggregate the results across neighboring sequences. As a result, they simultaneously process information spatially and temporally.

We now describe the operation that is performed in each STC layer. The notation below assumes that the input to layer $l$ is $h^{l-1} \in \reals^{T^{l-1} \times p \times c^{l-1}}$, where $T^{l-1}$ is the number of time points at the previous layer, and $h^{l-1}_i \in \reals^{T^{l-1} \times c^{l-1}}$ represents the hidden features associated with sequence $i$. We consider two propagation rules for STC layers, and in our experiments investigate which is most effective.

Both rules begin by applying a 1D convolution (denoted by $*$) with filter $W^l_{int}$ to each sequence of hidden features $h^{l-1}_i$, resulting in an intermediate set of features denoted by $a^l_i$. Note that filter $W^l_{int}$ is shared across all sequences in the layer $l$. The two rules differ in how they handle the aggregation of features from a node and its neighbors. Rule A aggregates features from the node's neighborhood, including the node itself, and then applies a nonlinearity $g$. The aggregation operation (e.g. mean, max) is performed along the spatial dimension, so that the temporal and channel dimensions are retained. Rule B first aggregates features across a node's neighbors, and then combines these features with the node's own features by concatenating them and passing them through an additional nonlinear operation, parameterized by $W^l_{comb}$. This prevents the node's feature from being ``diluted'' by the features from its neighboring nodes. Note that the distinction between these two propagation rules is similar to the distinction between rules proposed for graph convolutional network (GCN) \citep{kipf2016semi} and GraphSage \citep{hamilton2017inductive}.

The two rules for computing features for node $i$ at layer $l$ are:

\begin{align}
    &\mbox{\textbf{Rule A}} \\
    a^l_i &= W^l_{int} * h^{l-1}_i \nonumber \\
    z^l_i &= \text{AGGREGATE}(\{a^l_j \; \text{for} \; j \; \text{in} \; N(i)\}) \nonumber \\
    h^l_i &= g(z^l_i) \nonumber \\
    \nonumber \\
    &\mbox{\textbf{Rule B}} \\
    a^l_i &= W^l_{int} * h^{l-1}_i \nonumber \\
    z^l_i &= \text{AGGREGATE}(\{a^l_j \; \text{for} \; j \; \text{in} \; N(i) \setminus \; i \}) \nonumber \\
    h^l_i &= g_2\big( \: W^l_{comb} * g_1([z^l_i, \: a^l_i]) \: \big) \nonumber
\end{align} \\[1\baselineskip]

The neighborhood of node $i$ is defined as $N(i) = \{j \; \text{s.t.} \; A_{ij} = 1\}$, i.e. it's the set of nodes that have an edge to $i$. The only parameter for rule A is the convolutional kernel $W^l_{int} \in \reals^{t^l \times c^l \times c^{l-1}}$. The two parameters for rule B are $W^l_{int}$, and the second convolutional kernel $W^l_{comb} \in \reals^{t_2^l \times c^l \times (2*c^l)}$. In rule A the hyperparameters are the choice of nonlinearity $g$, the temporal kernel size $t^l$, and the number of channels $c^l$. Rule B has the additional hyperparameter $t^l_2$, which could simply be set to 1 or $t^l$, as well as the possibility of a second nonlinearity. For the two rules, note that filters are shared both spatially and temporally.

The adjacency matrix of the sequences is used when we aggregate features across $N(i)$. An interesting property of STC layers is the independence between the number of parameters and the input adjacency matrix, which allows a TGCN model to accept inputs with arbitrary graph topologies. This property is also critical for the investigation in section \ref{sec:explain} of dropping leads to interpret model predictions.

\subsection{\texorpdfstring{$k$}{k}-step reachable spatio-temporal convolutional layers}

The proposed STC layer aggregates information from nodes with a direct edge to the node it is operating on. We propose utilizing the concept of $k$-step reachability matrices to further incorporate information from nearby nodes that are reachable within $k$ steps. The $k$-step reachability matrix $A(k)$ is a binary matrix, which indicates nodes that can reach one another in at most $k$ steps. To obtain it from $A$, we use the operation $A(k) = \mathbbm{1}(A^k)$ where $A^k$ is the adjacency matrix raised to the $k$th power, $\mathbbm{1}(\cdot)$ is an element-wise indicator function, and $A(0) = I$. Setting $k > 1$ enables information to spread through the graph using fewer layers; setting $k = 0$ creates an STC layer that operates on each sequence separately. We incorporate $k$ as an additional hyperparameter for each STC layer. To reflect this in the STC propagation rules described above, $N(i)$ should be replaced by $N_k(i) = \{j \; \text{s.t.} \; A(k)_{ij} = 1\}$.

\subsection{Pooling layers}

For TGCN, we also explore the possibility of using pooling (max or mean) along the temporal dimension. This enables the model to leverage information across a wider time window without substantially increasing the number of layers. Similarly to standard CNNs for computer vision, TGCNs can have alternating STC layers and pooling layers.

\begin{figure}[!t]
  \centering 
  \includegraphics[width=\textwidth]{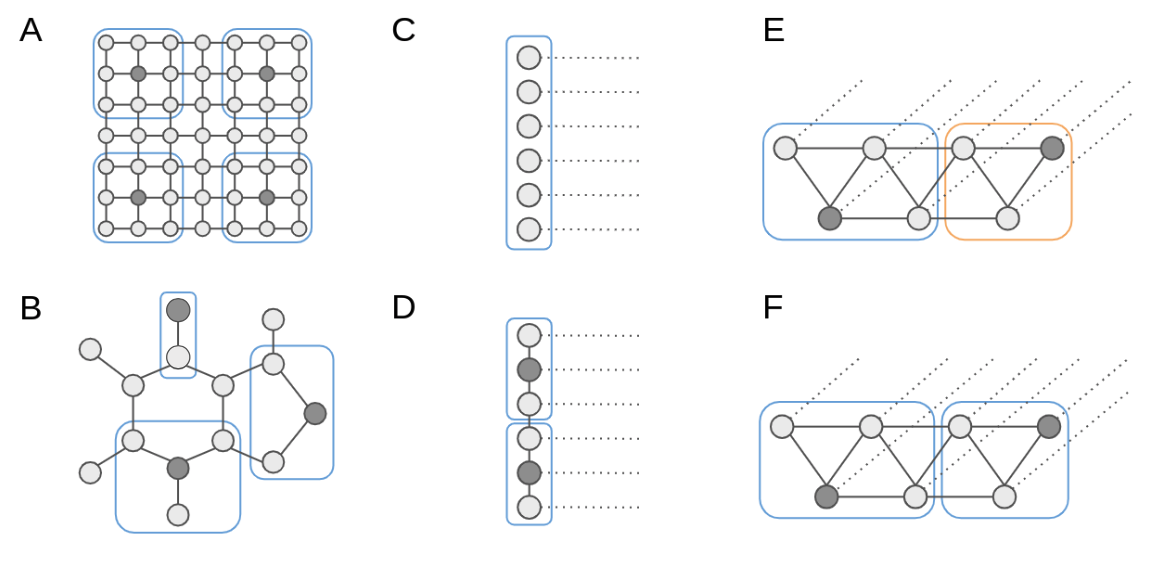}
  \caption{Comparison of localized and shared feature extraction operations across different models. Example inputs are shown for each model type. Receptive fields for one or more feature extraction operations are depicted as bounding boxes, where different colors indicate different sets of parameters. Dark colored nodes are those whose neighborhoods are being processed. \textbf{A:} CNN has feature extractors that are localized and shared over both spatial dimensions. \textbf{B:} GCN operates on arbitrary graphs, and has localized and shared feature extractors. \textbf{C:} TCNN has feature extractors that are localized and shared over time, but are not spatially localized. \textbf{D:} 2D TCNN has feature extractors that are localized and shared over time. They are also localized and shared over space, but spatial information is lost when nodes are put into a single ordering. \textbf{E:} SCNN has feature extractors that are localized and shared over time, but not shared spatially. \textbf{F:} TGCN has feature extractors that are localized and shared over both time and space.}
  \label{fig:concept} 
\end{figure}

\section{Prior work}

\subsection{Related models}

In this section, we discuss convolutional models for time series and neural network models for graph data, as both have strong connections with TGCN. We also highlight how each model leverages structural information, because it is challenging to utilize for non lattice-like graphs.

As the most generic convolutional model for time series, TCNN applies 1D convolutions over time at each layer \citep{lea2017temporal}. In doing so it combines information from all sequences in the first hidden layer without explicitly modeling the interactions between sequences. An extension that incorporates spatial information into TCNN is to apply 2D convolutions to subsets of neighboring sequences \citep{golmohammadi2017gated}. We term this model 2D TCNN, as it's architecturally similar to a TCNN but with 2D convolutions; it effectively approximates the true graph structure by placing nodes in a single ordering, possibly with repetitions. The structural convolutional neural network (SCNN) was proposed specifically to leverage structural information in time series \citep{teh2018generalised}.
It performs spatially localized feature extraction by learning a separate set of filters for operating on each node's neighborhood.

We also mention GNNs \citep{kipf2016semi, hamilton2017inductive, xu2018powerful} because of their connection with TGCNs. GNNs operate on data where each input example is an arbitrarily structured graph, and each node has a vector of features. GNNs use neighborhood aggregation schemes to gradually integrate information from nearby nodes, and thereby learns node representations on arbitrary graph data. Many GNN variants, such as GCNs \citep{kipf2016semi} and GraphSage \citep{hamilton2017inductive}, involve an aggregation step followed by a combination step: in each layer an operation is performed independently on each node, the results are aggregated across nodes' neighborhoods, and the aggregated features are then combined with the node's own features. TGCN requires a similar neighborhood aggregation scheme to handle the adjacency information in structural time series.

Figure \ref{fig:concept} provides graphical depictions of the models discussed above. The primary differences across the models are the input data structure and the degree of localized and shared feature extraction. Through this comparison, we show that TGCN is the only model that performs localized and shared feature extraction for structural time series, making it an analogue to CNNs for lattice-like graphs and GNNs for arbitrary graphs.

\subsection{Automatic seizure detection}

Many researchers have developed seizure detection algorithms in the last decade \citep{ahammad2014detection, furbass2015prospective, golmohammadi2017gated, golmohammadi2017deep, shoeb2010application, thodoroff2016learning, williamson2012seizure}. The task has typically been tackled as a per-epoch classification problem. Epochs in the literature range from sub-second to tens of seconds, and features (e.g. entropy, Fourier transform coefficients, wavelets, learned features, etc.) are computed from either the current epoch or neighboring epochs and used to classify the particular epoch of interest as seizure or non-seizure. The most accurate algorithms in the literature use traditional feature extraction and leverage patient-adaptive techniques, e.g. by using patient-specific thresholds \citep{furbass2015prospective, shoeb2010application}. Our work, like several other prominent studies, addresses the problem of cross-patient seizure detection \citep{golmohammadi2017gated, golmohammadi2017deep, thodoroff2016learning, wilson2004seizure}.

Modern deep learning approaches have also been explored for automatic seizure detection, and typically make efforts to utilize the structural information of EEGs \citep{golmohammadi2017gated, golmohammadi2017deep, thodoroff2016learning}. EEG montaging is a technique used by clinicians, which involves analyzing logical configurations of the EEG leads to aid localization and lateralization of EEG activities; some studies have therefore preprocessed the data by taking the difference between neighboring pairs of leads \citep{shoeb2010application}. However, all leads have multiple neighbors, so there are numerous montages to choose from, and each one captures spatial information imperfectly. One study integrated CNNs and RNNs in a model that used 2D convolutions to capture spatial and temporal information simultaneously, an idea that we refer to as 2D TCNN above \citep{golmohammadi2017gated}. Another study pre-processed data to create an image-like representation of EEGs, thereby overcoming the complex graph structures of the EEG leads and enabling the use of standard 2D CNNs \citep{thodoroff2016learning}; the image-like representation of EEGs was generated by interpolating a $16 \times 16$ image of the measurement from each lead.

\begin{table}[htbp]
  \centering 
  \caption{Dataset summary.} 
  \begin{tabular}{|l|r|r|r|}\hline
     & Number of patients & Total hours & Number of seizures\\ \hline
    Training & 995 & 14,741 & 18,436\\ \hline
    Tuning & 30 & 2,831 & 134\\ \hline
    Test & 38 & 4,083 & 171\\ \hline \hline
    Total & 1,063 & 21,655 & 18,741\\ \hline
  \end{tabular}
  \label{tab:data}
\end{table}

\section{Experiments}

\subsection{Data collection}

Scalp EEG data for this study was acquired from patients who underwent epilepsy diagnosis at the adult epilepsy monitoring unit (EMU) of
the Cleveland Clinic Foundation
between 2005 and 2017. EEG data was collected using standard 10-20 montage using 21 leads (see Appendix for the list of leads) and sampled at 200 Hz. Clipped EEG data consisting of baseline and seizure segments were acquired from 656 patients. Long term and continuous EEG data spanning the entire duration of EMU stay (weeks of recordings) was acquired from 398 patients. Temporal location of epileptic events such as seizures, interictal abnormalities were marked by a trained epileptologist. IRB approval was obtained prior to study initiation.

While both clipped and long term EEG data was used for training the algorithm, only long term EEG data was used for testing. Table \ref{tab:data} summarizes the contents of our dataset. A total of 995 patients were used for training the models, where clipped data came from 613 patients and continuous data came from 382 patients. The models were tested on long term EEG data from a hold-out test set of 38 patients. In total, we used approximately 15,000 hours of data for training and 4,000 hours for testing. Hyperparameter tuning was done on an independent tuning dataset of 30 patients with a total of 2,800 hours. There is no patient overlap across the training, tuning, and test datasets. Due to the low number of seizure events (positives), we subsampled the training dataset by filtering out 90\% of the negative examples.

\subsection{Data pre-processing}

In order to have uniformly sized inputs for our models, sessions were split into 96 second epochs. The per-epoch classification approach is common, but this window duration is relatively long. Our choice of 96 second windows was motivated by the fact that the average seizure length is around 50 seconds; in a clinical setting, such a system could be run every couple seconds, and the slightly longer history should be beneficial to the model. Each epoch's label was determined using clinician annotations, with positive examples being those epochs in which a seizure begins. This prediction target was chosen because clinicians are primarily interested in being notified when a seizure starts.

As a pre-processing step for the model, the EEG samples of size 19,200 x 21 were transformed into the frequency domain using the short time Fourier transform (STFT) with windows of size 64 and with overlap 32. Frequency transformations like STFT are common for high-frequency time series, in the domain of EEGs \citep{ahammad2014detection, furbass2015prospective, shoeb2010application, williamson2012seizure} but also others such as speech processing. Epileptologists analyze EEGs through their frequency information because seizures are known to involve electrical activity in a particular frequency band, so we expect the STFT to extract informative features. After performing the STFT, we retain the log magnitudes and omit the phase information, resulting in a tensor of size $599 \times 21 \times 33$.

\begin{table}[htbp]
  \centering 
  \caption{TGCN architecture configurations. Models with each of the following architectures are trained and compared to one another. Convolutions are performed with padding to retain the size along temporal dimension. STC $k$-$t$-$c$ denotes an STC layer with spatial kernel size $k$, temporal kernel size $t$, and $c$ channels.} 
  \begin{tabular}{|c|c|c|c|c|c|}\hline
    \multirow{2}{*}{} & \multicolumn{5}{c|}{TGCN Architectures}\\  \cline{2-6}
     & I & II & III & IV & V\\ \hline
    Raw input & \multicolumn{5}{c|}{Raw EEG (19,200 $\times$ 21)}\\ \hline
    Pre-processing & \multicolumn{5}{c|}{STFT (599 $\times$ 21 $\times$ 33)}\\ \hline
     \multirow{2}{*}{\centering Block 1} & STC 1-3-32 & \makecell{STC 1-3-32\\STC 1-3-32} & \makecell{STC 1-3-32\\STC 1-3-32\\STC 1-3-32} & \makecell{STC 0-3-32\\STC 1-3-32\\STC 1-3-32} & \makecell{STC 0-3-32\\STC 0-3-32\\STC 1-3-32\\STC 1-3-32}\\\cline{2-6}
     & \multicolumn{5}{c|}{max pooling along temporal dimension (300 $\times$ 21 $\times$ 32)}\\ \hline
     \multirow{2}{*}{\centering Block 2} & STC 1-3-64 & \makecell{STC 1-3-64\\STC 1-3-64} & \makecell{STC 1-3-64\\STC 1-3-64\\STC 1-3-64} & \makecell{STC 0-3-64\\STC 1-3-64\\STC 1-3-64} & \makecell{STC 0-3-64\\STC 0-3-64\\STC 1-3-64\\STC 1-3-64}\\\cline{2-6}
     & \multicolumn{5}{c|}{max pooling along temporal dimension (150 $\times$ 21 $\times$ 64)}\\ \hline
     \multirow{2}{*}{\centering Block 3} & STC 1-3-128 & \makecell{STC 1-3-128\\STC 1-3-128} & \makecell{STC 1-3-128\\STC 1-3-128\\STC 1-3-128} & \makecell{STC 0-3-128\\STC 1-3-128\\STC 1-3-128} & \makecell{STC 0-3-128\\STC 0-3-128\\STC 1-3-128\\STC 1-3-128}\\\cline{2-6}
     & \multicolumn{5}{c|}{max pooling along temporal dimension (75 $\times$ 21 $\times$ 128)}\\ \hline
     \multirow{2}{*}{\centering Block 4} & STC 1-3-256 & \makecell{STC 1-3-256\\STC 1-3-256} & \makecell{STC 1-3-256\\STC 1-3-256\\STC 1-3-256} & \makecell{STC 0-3-256\\STC 1-3-256\\STC 1-3-256} & \makecell{STC 0-3-256\\STC 0-3-256\\STC 1-3-256\\STC 1-3-256}\\\cline{2-6}
     & \multicolumn{5}{c|}{max pooling along temporal dimension (38 $\times$ 21 $\times$ 256)}\\ \hline
    \multirow{6}{3cm}{\centering For producing scalar prediction} & \multicolumn{5}{c|}{average pooling along spatial dimension (38 $\times$ 256)}\\ \cline{2-6}
     & \multicolumn{5}{c|}{flatten (9,728 units)}\\ \cline{2-6}
     & \multicolumn{5}{c|}{fully connected 512}\\ \cline{2-6}
     & \multicolumn{5}{c|}{fully connected 512}\\ \cline{2-6}
     & \multicolumn{5}{c|}{fully connected 1}\\ \cline{2-6}
     & \multicolumn{5}{c|}{sigmoid}\\ \hline
  \end{tabular}
  \label{tab:configs}
\end{table}

\subsection{Architecture exploration}

In order to understand the role of the various hyperparameters involved in a TGCN's architecture, we compare five different configurations (Table \ref{tab:configs}). There is a large hyperparameter search space, but these configurations give a sense of the importance of some hyperparameter choices.

Each model has four blocks that consist of one or more STC layers, and are followed by a pooling layer.
Nonlinearities are preceded by batch normalization \citep{ioffe2015batch}, and the two fully connected layers use dropout \citep{srivastava2014dropout} with probability 0.2. We train the models using cross-entropy loss, stochastic gradient descent with momentum, and decay the learning rate after every 100 gradient steps. We use the ReLU nonlinearity in all models. We explore each configuration with both propagation rules (A and B) described in section \ref{sec:math}. For both rules we aggregate features across node's neighborhoods using the max operation, and for rule B we set $t_2^l = 1$ in all STC layers.

When characterizing model performance, we are motivated by the use case of an alarm system, where clinicians are notified when a patient has a seizure. We collect five metrics: area under the receiver operating curve (AU-ROC), area under the precision-recall curve (AU-PR), F1, sensitivity at 97\% specificity, and sensitivity at 99\% specificity. The first three metrics are standard for measuring performance in classification tasks, but for an alarm system in a clinical setting, the last two metrics are most relevant. A low false alarm rate is critically important for clinicians to rely on such a tool, because it's infeasible to respond to a tool that creates too many distracting false alarms.

Each of the models were trained on the training set and evaluated on the tuning set (Table \ref{tab:data}). Due to the time required to train models, only one of each configuration was trained. The experiment results are summarized in Table \ref{tab:tuning_comps}. The results show that 
the best performance is achieved by rule B and configuration II, but that there isn't a consistent pattern in one rule outperforming the other, or in additional layers improving accuracy.
This may be due to noisy results arising from our relatively small tuning set.
For simplicity, due to the strong performance of configuration II and propagation rule B, we use this model in subsequent experiments.

\begin{table}[htbp]
  \centering 
  \caption{TGCN configurations on the tuning set. The recorded performance is the best level achieved during training. Results are obtained from single runs. The best performance according to each metric is bolded.} 
  \begin{tabular}{|c|c|c|c|c|c|c|}\hline
    Config. & Prop. rule & AU-ROC & AU-PR & F1 & Sens.@97\% Spec. & Sens.@99\% Spec. \\ \hline
    \multirow{2}{*}{\centering I} & A & 0.965 & 0.289 & 0.376 & 0.842 & 0.684 \\ \cline{2-7}
     & B & \textbf{0.971} & 0.265 & 0.351 & 0.805 & 0.729 \\ \hline
    \multirow{2}{*}{\centering II} & A & 0.966 & 0.308 & 0.433 & 0.835 & 0.729 \\ \cline{2-7}
     & B & 0.969 & \textbf{0.364} & \textbf{0.450} & \textbf{0.872} & \textbf{0.752} \\ \hline
    \multirow{2}{*}{\centering III} & A & 0.970 & 0.268 & 0.348 & 0.827 & 0.729 \\ \cline{2-7}
     & B & 0.959 & 0.273 & 0.358 & 0.812 & 0.722 \\ \hline
    \multirow{2}{*}{\centering IV} & A & 0.967 & 0.316 & 0.432 & 0.842 & 0.737 \\ \cline{2-7}
     & B & 0.957 & 0.177 & 0.265 & 0.797 & 0.654 \\ \hline
    \multirow{2}{*}{\centering V} & A & 0.960 & 0.268 & 0.352 & 0.820 & 0.699 \\ \cline{2-7}
     & B & 0.957 & 0.258 & 0.348 & 0.805 & 0.699 \\ \hline
  \end{tabular}
  \label{tab:tuning_comps}
\end{table}

\subsection{Comparison with related models}

In section 4.2 we experimented with some of the hyperparameters that lead to a TGCN with strong performance. Here, we put TGCN's performance in context through comparisons with several related models that take different approaches to leveraging structural information.

\textbf{TCNN} applies convolutions along the temporal dimension, and neglects to utilize structural information. \textbf{TCNN with montage} is a TCNN that operates on differenced sensor data instead of raw sensor data (see Appendix for details). \textbf{2D TCNN} approximates the graph structure using a single ordering, setting up a list of leads and applying 2D convolutions (see Appendix for details), as in \citep{golmohammadi2017gated}. \textbf{2D TCNN with montage} is a 2D TCNN that operates on differenced sensor data (see Appendix for details). \textbf{SCNN} has separate filters for operating on each node's neighborhood, making it well suited for structural time series \citep{teh2018generalised}. In order to have comparable architectures, these models were trained with four blocks of convolutional layers, using the same number of channels as the TGCNs in the previous experiments. We performed identical preprocessing with the STFT, applied similar optimization-related hyperparameters, and used the tuning set to determine the best number of layers.

Each of the models were trained 10 times, and the best model from each run (chosen according to the tuning set) was run on the test data. We characterize the performance of these models by the mean and standard deviation of their performance. Table \ref{tab:test_comps} shows the results of these experiments. 2D TCNN appears to have the highest mean sensitivity at 99\% specificity, whereas SCNN seems to have the highest mean sensitivity at 97\% specificity. TGCN's mean performance is roughly in the middle for AU-ROC and sensitivity at 99\% specificity, but on the lower end for sensitivity at 97\% specificity. However, there is relatively high variance between runs, so that when taking the standard deviations into account, the models appear to perform quite similarly.


There are several possible methods for improving the performance of TGCN (as well as the baseline models) that we did not explore. For simplicity, we did not explore integrating RNNs as in \citep{golmohammadi2017deep, thodoroff2016learning}, adding residual connections, or using other techniques for managing the large class imbalance. We also emphasize that it's not possible to draw a direct comparison between these performance metrics and those presented in other studies, because of the strong dependence on the cohort in the test set. However, since the primary goal was to observe the relative difference in performance between TGCN and other models, these factors should not alter the conclusions.

\begin{table}[htbp]
  \centering 
  \caption{Comparison of TGCN with baseline models. Each entry shows the mean and standard deviation across 10 runs.} 
  \begin{tabular}{|l|c|c|c|}\hline
     & AU-ROC & Sens.@97\% Spec. & Sens.@99\% Spec.\\ \hline
    TCNN & $0.926 \pm 0.019$ & $0.648 \pm 0.042$ & $0.466 \pm 0.034$\\ \hline
    TCNN montage & $0.935 \pm 0.006$ & $0.645 \pm 0.059$ & $0.455 \pm 0.054$\\ \hline
    2D TCNN & $0.927 \pm 0.011$ & $0.645 \pm 0.033$ & $0.485 \pm 0.042$\\ \hline
    2D TCNN montage & $0.917 \pm 0.026$ & $0.653 \pm 0.089$ & $0.470 \pm 0.074$\\ \hline
    SCNN & $0.931 \pm 0.031$ & $0.669 \pm 0.061$ & $0.460 \pm 0.039$\\ \hline
    TGCN & $0.928 \pm 0.008$ & $0.635 \pm 0.022$ & $0.467 \pm 0.038$\\ \hline
  \end{tabular}
  \label{tab:test_comps}
\end{table}

\section{Model explainability} \label{sec:explain}

Analyzing EEG is a time-consuming task that can only be performed by expert readers. The classification models developed in the previous section provide a way of accelerating this process by identifying 96 second windows when seizures likely occurred. However, an important obstacle to the adoption of deep learning in healthcare applications is the lack of transparency into how models make decisions. Techniques for explaining model predictions can be useful to help establish trust with clinicians, to better understand the model's shortcomings, and even for generating new clinical insights \citep{ting2017development}.

In this section we explore two explainability approaches for TGCN, and demonstrate that we can extract rich contextual information from the windows when seizures occurred. Specifically, we can help clinicians determine when precisely the seizure occurred, and the part of the brain that was most involved. To demonstrate this approach, we show examples where we ask an epileptologist to look at the same EEGs that were analyzed by the algorithm, and give their clinical diagnosis.

The dominant paradigm for interpreting deep learning models is input attribution, which explains a prediction by assigning an importance level to each of the input features. Gradient backpropagation-based methods are most common \citep{smilkov2017smoothgrad, sundararajan2017axiomatic, simonyan2013deep}, and they're applicable here because TGCNs are fully differentiable, and the STFT operation is differentiable as well. We calculate gradients of the model logit with respect to the raw waveforms, and average the gradients across five models; we find that this leads to better visualizations than a single model, because single models tend to highlight only a subset of salient parts of the input. Since the sign of the gradient is of no importance, we take the absolute value after averaging.

\begin{figure}[htbp]
  \centering 
  \includegraphics[width=\textwidth]{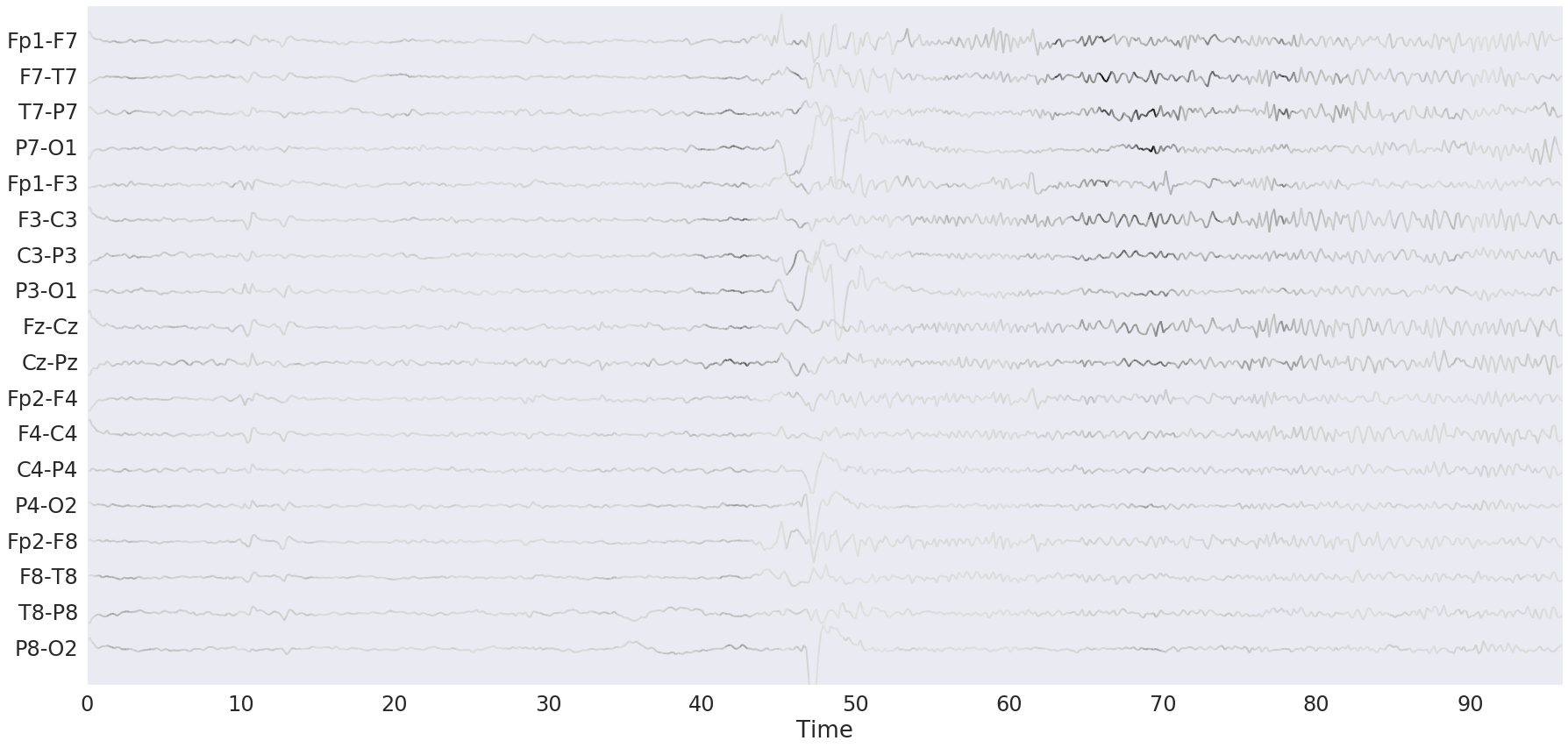}
  \caption{EEG attribution overlay plot. The EEG sample is visualized using the 10-20 montage, and with the waveform intensity proportional to the attribution score.}
  \label{fig:overlay} 
\end{figure}

Figure \ref{fig:overlay} shows an EEG sample with the attribution scores overlaid. It depicts a 96 second window where a seizure occurred, and the intensity of the waveforms indicates which parts of the EEG were most influential on the model's prediction. The EEGs are visualized using the 10-20 montage, so we also add the attribution scores for pairs of leads that are differenced.

A clinician who evaluated this seizure remarked that the patient was initially asleep, until higher frequency EEG and muscle artifacts from patient arousal occurred at around 40-45 seconds into the sample. The high frequencies are most prominent in the left part of the brain, and evolve into high amplitude sharp waves; they're present primarily in the Fp1, F7 and F3 leads, and proceed for the remainder of the sample. The visualization corroborates this description, especially through its emphasis on the activity beginning at 40 seconds, and the subsequent emphasis on high frequency activity in the left part of the brain (the top half of Figure \ref{fig:overlay}).

Our second approach for interpreting TGCNs relies on their capacity to accept inputs with different graph topologies. We investigate whether it's possible to perform spatial localization of seizures in the brain by ignoring sets of leads, and observing how their omission impacts the model's prediction. We call this method \textit{sequence dropout}. Two methods for performing sequence dropout are ignoring one lead at a time, and simultaneously ignoring multiple leads from one part of the brain. Sequence dropout is similar to the model-agnostic technique of input occlusion \citep{zeiler2014visualizing}, except that examples modified by zeroing leads entirely are less likely to resemble realistic EEG samples. The ability to use this method is an advantage for TGCN, arising from its unique method of processing structural time series.

\begin{figure}[t!]
  \centering 
  \includegraphics[width=0.8\textwidth]{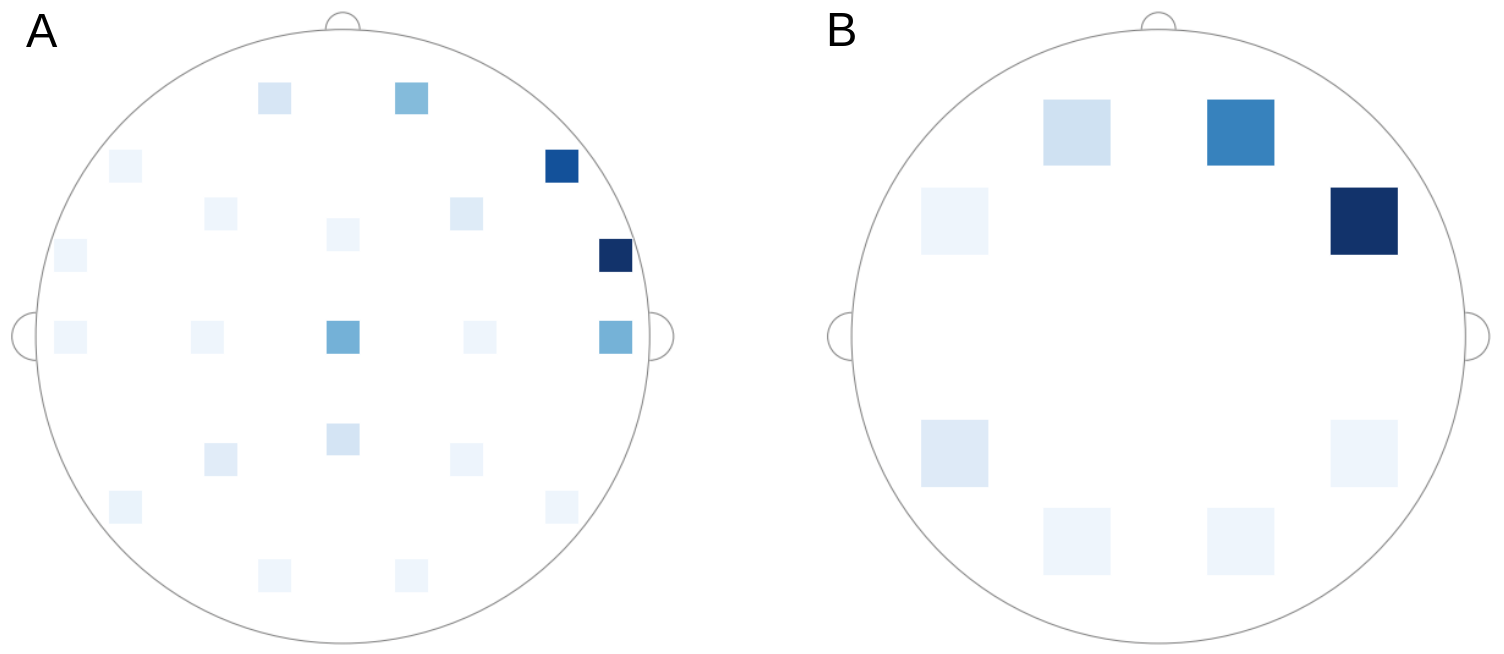}
  \caption{Seizure localization with sequence dropout. The intensity shown is determined by the reduction in the model's prediction when the corresponding leads were ignored. \textbf{A:} dropping one lead at a time. \textbf{B:} dropping sets of leads that correspond to different regions of the brain.}
  \label{fig:dropout} 
\end{figure}

Sequence dropout is likely to provide the most insight for focal seizures, i.e. those that are confined to a small region of the brain. Figure \ref{fig:dropout} shows an example of sequence dropout applied to a focal seizure. In our visualization, the intensity at each location is determined by the reduction in the model's logit when the corresponding lead or set of leads is omitted from the input. (For the particular sets of leads that were used in Figure \ref{fig:dropout}B, see the Appendix.) Similarly to the above method, we run sequence dropout using an ensemble of models.

A clinician who analyzed the seizure depicted in Figure \ref{fig:dropout} noted that it occurred over the right part of the brain, and is particularly visible in the leads F8, F4, and T8. The visualization in Figure \ref{fig:dropout}A clearly places emphasis on the leads in the right temporal part of the brain, especially F8, T8 and FT10, corroborating the clinician's description. Figure \ref{fig:dropout}B also gives high attribution to the right side of the brain, primarily in the right temporal region, but also to some extent in the right frontal region.

Both of the model explainability techniques reveal rich information that was not available during training. Model inputs that were 96 seconds in length were marked as positive or negative, but the model was never given precise timing or localization of seizures. These model attribution techniques enable finer temporal and spatial localization of seizures, which may be regarded as semi-supervised tasks.

Due to the difficulty of these problems, both techniques have shortcomings. The gradient-based method is limited by the fact that it does not highlight all salient information in the input; since the score is based on relative gradient magnitudes, some parts of the input that contain evidence for a seizure may nonetheless receive low attribution. For the sequence dropout method, even when leads that are most involved in a seizure are ignored, activity in the other leads (e.g. muscle artifact that is characteristic of a seizure) can be strongly indicative that a seizure is occurring. Furthermore, both attribution approaches confront the obstacle that it's fundamentally unclear how to ascribe importance to different parts of an EEG sample.

\section{Conclusion}

In this paper we proposed the temporal graph convolutional network (TGCN), a deep learning model for structural time series. Unlike other approaches for analyzing time series data, TGCN applies feature extraction operations that are shared over both time and space. Localized and shared feature extraction enforces a useful invariance for certain applications, and provides a useful inductive bias for our task of automatic seizure detection from EEGs. In our experiments, we showed that deep learning models can achieve strong performance in automated seizure detection, and that TGCN's performance matches that of several related models.

Our attribution visualizations provide novel approaches to better understand how deep learning models perform seizure detection on EEG data. They shed light into the ``black box'' that is a deep learning model, suggesting that our models focus on similar features of EEGs as clinicians do. Deep learning-based seizure detection algorithms, along with the model explainability techniques that we presented, could significantly aid clinicians in accelerating the labor-intensive and time-consuming process of interpreting EEGs and diagnosing epilepsy. 

The methods developed in this work could also be applied to other structural time series data. TGCN unites ideas from graph neural networks (GNNs) and convolutional models for time series, making it well suited for graph-structured time series. These types of time series data exist elsewhere in medicine, e.g. in multi-lead electrical signals from the heart such as EKGs, sleep study data such as PSGs, and pressure waveforms from intrathoracic catheters.

TGCN is particularly promising for problems where multiple datasets with similar but not identical configurations needs to be pooled, such as if one wants to train a model using a pooled dataset from hospitals that used different lead setups in EEG sessions. It may also prove to be less prone to overfitting when applied to smaller datasets, where memorization of the training set is a larger concern.

\acks{We would like to thank Justin Tansuwan, Rebecca Davies, Hector Yee, Jiang Wu, and Dasarathi Sampath for their valuable help in managing the data, creating visualization tools, and discussing modeling ideas. We would also like to thank Cameron Chen for his valuable help in writing the manuscript.}


\clearpage
\appendix
\section*{Appendix}

\begin{table}[htbp]
\centering 
  \caption{Set of EEG leads analyzed, and how they were arranged for several models.}
\begin{tabular}{|l|l|l|l|}
\hline
\multicolumn{1}{|c|}{Leads analyzed}                                                                                                                & \multicolumn{1}{c|}{\begin{tabular}[c]{@{}c@{}}Inputs for TCNN\\ with montage\end{tabular}}                                                                                                                                                                         & \multicolumn{1}{c|}{\begin{tabular}[c]{@{}c@{}}Input ordering for\\ 2D TCNN\end{tabular}}                                                                                                          & \multicolumn{1}{c|}{\begin{tabular}[c]{@{}c@{}}Input ordering for 2D\\ TCNN with montage\end{tabular}}                                                                                                                                                              \\ \hline
\begin{tabular}[t]{@{}l@{}}C3\\ C4\\ Cz\\ F3\\ F4\\ F7\\ F8\\ Fz\\ FT9\\ FT10\\ Fp1\\ Fp2\\ O1\\ O2\\ P3\\ P4\\ P7\\ P8\\ Pz\\ T7\\ T8\end{tabular} & \begin{tabular}[t]{@{}l@{}}Fp1 - F7\\ F7 - T7\\ T7 - P7\\ P7 - O1\\ \\ Fp1 - F3\\ F3 - C3\\ C3 - P3\\ P3 - O1\\ \\ Fz - Cz\\ Cz - Pz\\ \\ Fp2 - F4\\ F4 - C4\\ C4 - P4\\ P4 - O2\\ \\ Fp2 - F8\\ F8 - T8\\ T8 - P8\\ P8 - O2\\ \\ FT9 - F7\\ FT10 - F8\end{tabular} & \begin{tabular}[t]{@{}l@{}}Fp1\\ F7\\ T7\\ P7\\ O1\\ \\ Fp1\\ F3\\ C3\\ P3\\ O1\\ \\ Fz\\ Cz\\ Pz\\ \\ Fp2\\ F4\\ T4\\ P4\\ O2\\ \\ Fp2\\ F8\\ T8\\ P8\\ O2\\ \\ FT9\\ F7\\ FT10\\ F8\end{tabular} & \begin{tabular}[t]{@{}l@{}}Fp1 - F7\\ F7 - T7\\ T7 - P7\\ P7 - O1\\ \\ Fp1 - F3\\ F3 - C3\\ C3 - P3\\ P3 - O1\\ \\ Fz - Cz\\ Cz - Pz\\ \\ Fp2 - F4\\ F4 - C4\\ C4 - P4\\ P4 - O2\\ \\ Fp2 - F8\\ F8 - T8\\ T8 - P8\\ P8 - O2\\ \\ FT9 - F7\\ FT10 - F8\end{tabular} \\ \hline
\end{tabular}
\end{table}

\begin{table}[htbp]
  \centering 
  \caption{Lead groups for brain regions. These groups were used for generating Figure \ref{fig:dropout}B from the sequence dropout experiment.} 
  \begin{tabular}{|c|c|}\hline
     & Number of seizures\\ \hline
    Left frontal & Fp1, F3, Fz\\ \hline
    Right frontal & Fp2, F4, Fz\\ \hline
    Left temporal & F7, T7, FT9\\ \hline
    Right temporal & F8, T8, FT10\\ \hline
    Left parietal & P7, P3\\ \hline
    Right parietal & P8, P4\\ \hline
    Left occipital & O1, P7\\ \hline
    Right occipital & O2, P8\\ \hline
  \end{tabular}
\end{table}

\end{document}